\documentclass{llncs}
\renewcommand{\abstractname}{Abstract}
\usepackage{xspace}
\usepackage{graphicx}
\usepackage{booktabs}
\usepackage{array}
\usepackage{ragged2e}
\usepackage{ltablex}
\keepXColumns
\usepackage{makecell}
\usepackage{microtype}
\usepackage{xcolor}
\usepackage{placeins}
\usepackage{adjustbox}
\usepackage[hidelinks]{hyperref}
\usepackage{float}
\usepackage{caption}
\usepackage{pdfpages}

\usepackage{xeCJK}

\newcolumntype{L}{>{\RaggedRight\arraybackslash}X}
\newcolumntype{P}[1]{>{\RaggedRight\arraybackslash}p{#1}}

\renewcommand{\abstractname}{Abstract}
\renewcommand{\figurename}{Figure}
\renewcommand{\tablename}{Table}
\renewcommand{\refname}{References}
\titlerunning{Chinese-SkillSpan}
\authorrunning{Guojing Li et al.}

\begin{document}

\title{Chinese-SkillSpan: A Span-Level Dataset for ESCO-Aligned Competency Extraction from Chinese Job Ads}

\author{
\mbox{Guojing Li}\inst{1,2} \and
\mbox{Zichuan Fu}\inst{1} \and
\mbox{Junyi Li}\inst{1} \and
\mbox{Wenxia Zhou}\inst{2} \and
\mbox{Xinyang Wu}\inst{1} \\
\mbox{Jinning Yang}\inst{1} \and
\mbox{Jingtong Gao}\inst{1} \and
\mbox{Feng Huang}\inst{1} \and
\mbox{Xiangyu Zhao}\inst{1}
}

\institute{
City University of Hong Kong \and
Renmin University of China
}

\maketitle

\begingroup
\renewcommand\thefootnote{}
\footnotetext{Guojing Li and Zichuan Fu contributed equally to this work. Corresponding authors: Guojing Li (e-mail: guojingli3-c@my.cityu.edu.hk) and Xiangyu Zhao (e-mail: xianzhao@cityu.edu.hk).}
\endgroup

\begin{abstract}
Job Skill Named Entity Recognition (JobSkillNER) aims to automatically extract key skill information from large-scale job posting data, which is important for improving talent-market matching efficiency and supporting personalized employment services. To the best of our knowledge, this work presents the first Chinese JobSkillNER dataset for recruitment texts. We propose annotation guidelines tailored to Chinese job postings and an LLM-empowered Macro--Micro collaborative annotation pipeline. The pipeline leverages the contextual understanding ability of large language models (LLMs) for initial annotation and then refines the results through expert sentence-level adjudication. Using this pipeline, we annotate more than 20,000 instances collected from four major recruitment platforms over the period 2014--2025. Based on these efforts, we release Chinese-SkillSpan, the first Chinese JobSkillNER dataset aligned with the ESCO occupational skill standard across four dimensions: knowledge, skill, transversal competence, and language competence (LSKT). Experimental results show that the dataset supports effective model training and evaluation, indicating that Chinese-SkillSpan helps fill a major gap in Chinese JobSkillNER resources and provides a useful benchmark for intelligent recruitment research. Code and data are available at \url{https://sites.google.com/view/cn-skillspan-resources}.
\keywords{Chinese JobSkillNER \and skill extraction \and ESCO \and dataset construction \and large language models}
\end{abstract}

\section{Introduction}
\label{sec:intro}

With the rapid growth of online recruitment platforms, large volumes of job postings have become an important resource for understanding labor-market demand and supporting talent--position matching \cite{ILO_2020_OJVs_BigData}. A central task in this setting is Job Skill Named Entity Recognition (JobSkillNER), which aims to identify and extract textual spans that express job-related competencies, such as professional skills, knowledge requirements, language ability, and transversal competences, from unstructured job descriptions. Reliable JobSkillNER is a prerequisite for downstream applications such as skill profiling, job recommendation, and labor-market intelligence.

Recent work has begun to establish span-level, ontology-aware benchmarks for job-skill extraction in high-resource languages. In particular, English resources such as SkillSpan have shown the value of expert-annotated, ESCO-aligned span labels for benchmarking extraction models and studying boundary-sensitive competency recognition \cite{zhang2022skillspan}. However, despite the practical importance of Chinese recruitment data and the growing demand for skill intelligence in Chinese labor markets, there is still no publicly available Chinese span-level JobSkillNER dataset aligned with an international skill ontology. This resource gap limits both method development and cross-lingual comparison.

Constructing such a dataset is non-trivial. Large-scale manual annotation is costly and time-consuming, while recruitment texts contain substantial ambiguity in boundary selection, competency typing, and context dependence. These challenges are particularly salient in Chinese, where compressed expressions, implicit subjects, coordination structures, and domain-specific wording often make span boundaries harder to standardize. Recent studies suggest that LLM-assisted annotation can substantially improve efficiency and, when combined with human validation, approach high-quality labeling outcomes \cite{smallm,oliveira}. Nevertheless, existing work has not provided a Chinese-specific annotation framework for JobSkillNER that simultaneously addresses span consistency, ontology alignment, and scalable human--LLM collaboration.

To address these gaps, we introduce \textbf{Chinese-SkillSpan}, a Chinese span-level JobSkillNER dataset aligned with the ESCO standard \cite{ESCOv1_2_2024}. The dataset is constructed from four heterogeneous recruitment sources spanning 2014--2025, covering both domain and temporal diversity. We further develop an \textbf{LLM-empowered Macro--Micro collaborative annotation pipeline}, in which LLMs first generate context-aware draft annotations and human annotators then conduct sentence-level adjudication to produce the final gold labels. To improve consistency, we design annotation guidelines tailored to Chinese recruitment texts, including minimal-sufficient span boundaries, explicit type disambiguation rules, and offline ESCO alignment for version-stable ontology mapping. Annotation quality is assessed using inter-annotator agreement, including span-level F1 and Cohen's $\kappa$ \cite{landis1977measurement}.

Experimental results show that models fine-tuned on Chinese-SkillSpan achieve strong performance, indicating that the dataset provides a useful benchmark for Chinese JobSkillNER and a practical foundation for intelligent recruitment research. Overall, this work makes the following contributions:

\begin{itemize}
    \item We introduce \textbf{Chinese-SkillSpan}, a Chinese span-level JobSkillNER dataset aligned with ESCO, filling an important resource gap for Chinese recruitment text mining.
    \item We propose a \textbf{Macro--Micro collaborative annotation pipeline} that combines LLM-based draft annotation with expert adjudication, together with annotation guidelines tailored to Chinese recruitment texts.
    \item We provide initial benchmark results showing that Chinese-SkillSpan supports effective model training and evaluation for Chinese JobSkillNER.
\end{itemize}

\begin{figure*}[t]
  \centering
  \includegraphics[width=0.92\linewidth]{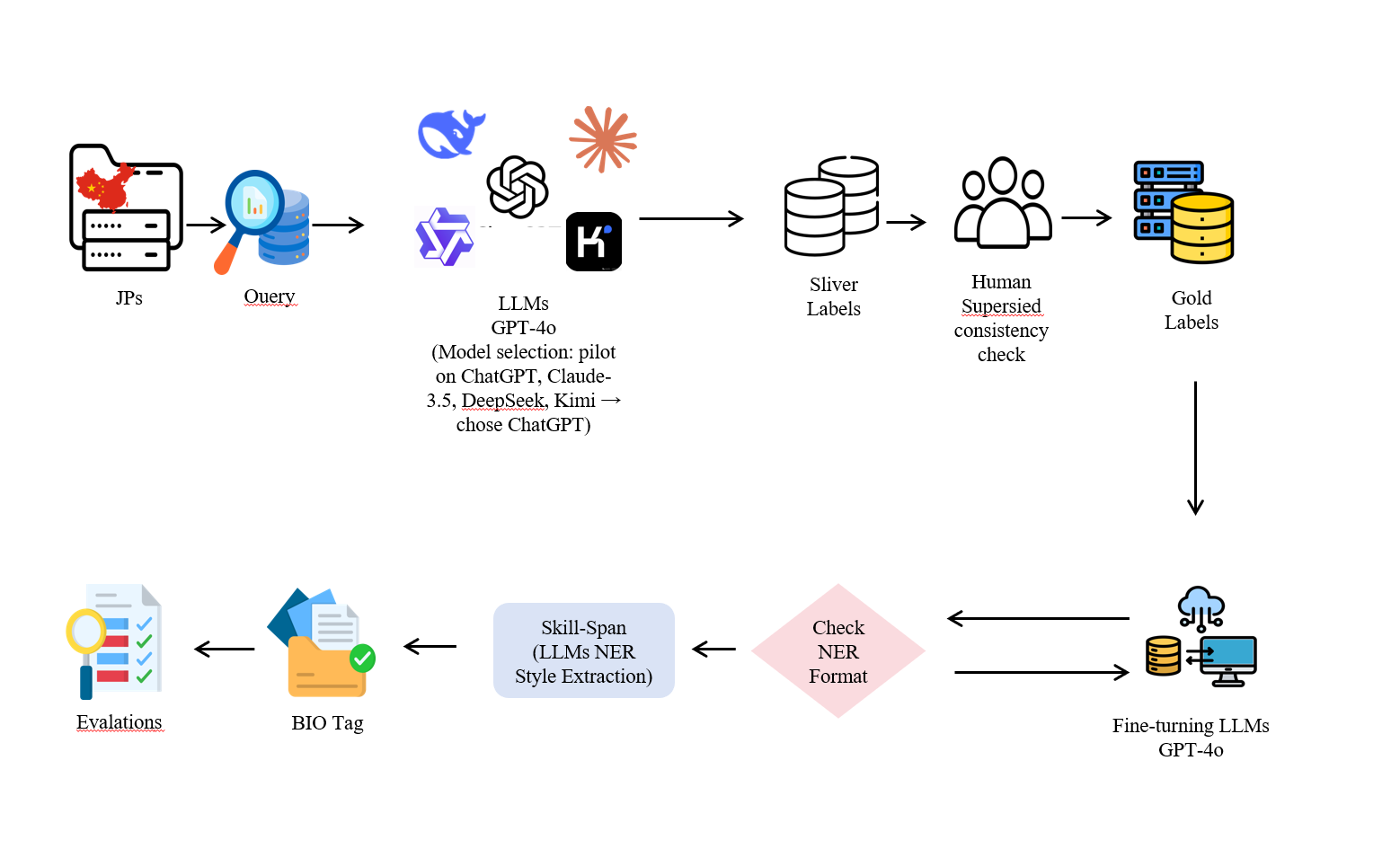}
  \caption{Overview of the \textbf{LLM-empowered Macro--Micro collaborative annotation pipeline}.}
  \label{fig:macro-micro}
  \vspace*{-3mm}
\end{figure*}

\section{Task Definition and Annotation Framework}
\label{sec:framework}

We present \textbf{Chinese-SkillSpan}, a span-level competency dataset for Chinese job ads with a minimal benchmark and a transparent extraction stack. Spans follow ESCO's \textbf{LSKT} taxonomy and are aligned to ESCO concept IDs. To better capture soft-skill nuances while keeping the top-level scheme intact, we optionally use local tags (A1--A3) for analysis without changing \textbf{LSKT}.
\newline
\textit{Labeling pipeline.} Four instruction-tuned LLMs generate draft spans and ESCO candidates; we aggregate by majority plus confidence and then conduct double/triple human adjudication in \textit{Doccano}, releasing \textbf{Silver} and \textbf{Gold} layers with agreement statistics (Cohen's~$\kappa$).
\newline
\textit{Our method (for the benchmark).} A reliability-focused stack: supervised fine-tuning of a Chinese LLM (e.g., Qwen) with span-anchored, dataset-specific prompting and constrained decoding, offline ESCO retrieval for definition grounding, a \emph{SelfCheck} module for boundary/ontology sanity checks, and a conditional reflection pass. A lightweight sentence classifier can gate decoding on negative sentences.
\newline
Evaluation uses a unified scorer with \textbf{Exact-span F1}, \textbf{Relaxed F1} (IoU$\ge$0.5), and \textbf{Concept Accuracy}. We release fixed IID and industry/time-shifted OOD splits, and baselines: ESCO lexicon match; supervised sequence labeling (BERT-CRF, span-based); cross-lingual zero-shot (XLM-R); and our LLM stack. All data, scripts, and a data card are released for reproducibility.

\begin{table}[t]
\caption{Corpus statistics for the \textsc{Chinese-SkillSpan} dataset.}
  \centering
  \footnotesize
  \setlength{\tabcolsep}{5.5pt}
  \renewcommand{\arraystretch}{1.12}
  \resizebox{\columnwidth}{!}{%
    \begin{tabular}{lrrrrrrr}
      \toprule
      \textbf{Split} & \textbf{\#Sent} & \textbf{Avg Len} & \textbf{Avg 4D} & \textbf{Avg L} & \textbf{Avg K} & \textbf{Avg S} & \textbf{Avg T} \\
      \midrule
      dev   & 2{,}143  & 40.37 & 3.607 & 0.016 & 1.276 & 1.810 & 0.504 \\
      test  & 3{,}237  & 43.85 & 2.306 & 0.010 & 0.822 & 1.141 & 0.332 \\
      train & 17{,}460 & 37.41 & 2.354 & 0.019 & 0.639 & 1.183 & 0.513 \\
      \bottomrule
    \end{tabular}%
  }
   \textit{Avg Len} is the average number of characters. \textit{Avg 4D} is the mean number of annotated skill spans per sentence. \textit{Avg L/K/S/T} breaks down Language, Knowledge, Specific, and Transversal skills.
  \label{tab:skillspan-eacl-stats}
  \vspace{-0.3cm}
\end{table}

\section{Experiments}
\label{sec:experiments}

\subsection{Data and Sampling}
We evaluate on the proposed \textbf{Chinese-SkillSpan}, which is constructed to reflect both source diversity and temporal variation in Chinese recruitment texts. The corpus is drawn from four heterogeneous recruitment streams: (i) AI/technology-oriented job boards at million scale, (ii) a graduate job portal at million scale, (iii) Alibaba open postings (approximately 25--30K), and (iv) public-sector recruitment texts (approximately 6.5K, with generally longer descriptions). To reduce source bias and temporal skew, we adopt \textbf{stratified random sampling} by source, year/quarter, and industry, covering the period from \textbf{2014 to 2025}. This design aims to improve representativeness and mitigate long-tail sparsity as well as domain/time drift.

\subsection{Annotation Protocol and Quality Control}
Annotation follows a \textbf{flat, non-nested LSKT} scheme with \emph{minimal-sufficient} span boundaries. Each instance is presented together with its \emph{job title} and a left/right \emph{context window} to provide situational cues for disambiguation. Two trained annotators label each instance independently, and a third senior annotator resolves disagreements when necessary. The annotation guideline was iteratively refined through human--LLM collaborative drafting, including boundary templates and conflict-resolution rules with a fallback hierarchy of $L > S > K > T$.

To improve scalability while preserving quality, we first construct \textbf{Silver} annotations from four LLMs through majority voting with confidence-based aggregation, and then derive the final \textbf{Gold} labels via double or triple human adjudication in \textit{Doccano}. The released scorer enforces both \emph{type} and \emph{boundary} legality under the flat LSKT scheme.

\begin{figure}[H]
    \centering
    \includegraphics[width=0.95\linewidth]{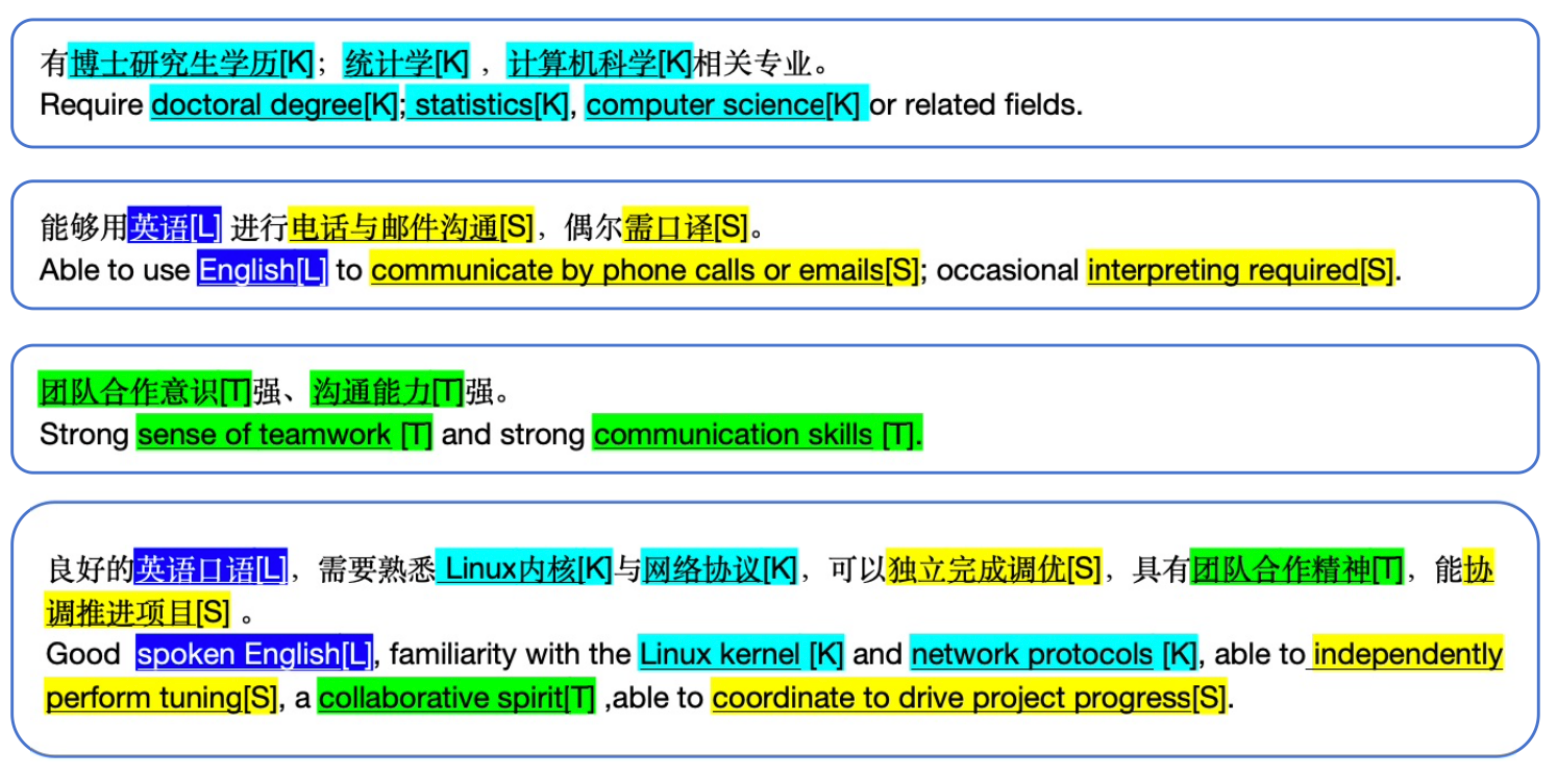}
    \caption{Representative examples illustrating span boundaries and label assignment under the flat, non-overlapping LSKT scheme for Chinese job advertisements.}
    \label{fig:sktl-sample}
\end{figure}

\subsection{Inter-Annotator Agreement}
To assess annotation reliability, we measure inter-annotator agreement (IAA) on a 100-sample gold subset annotated by two expert coders under the same LSKT guideline. Span-level agreement is computed under both \textit{strict} (exact boundary match) and \textit{relaxed} (overlapping boundary match) criteria, following prior span-level job-skill annotation work \cite{zhang2022skillspan}. For complementary assessment, we also report token-level Cohen's~$\kappa$.


As shown in Table~\ref{tab:iaa}, the annotation process achieves a strict span-level F1 of 0.532, a relaxed span-level F1 of 0.624, and a token-level Cohen's~$\kappa$ of 0.554. The gap between strict and relaxed agreement suggests that boundary decisions remain a major source of difficulty, especially for multi-word competency phrases. At the same time, the overall agreement level indicates a reasonably stable annotation process for this dataset construction setting.

\subsection{Evaluation Protocol}
We consider one \textbf{IID} setting and two \textbf{OOD} settings: \textit{Industry-OOD}, where selected industries are held out during training, and \textit{Time-OOD}, where a held-out quarter is used for temporal generalization testing. We report \textbf{Exact-span F1}, \textbf{Relaxed F1} (IoU $\ge 0.5$), and \textbf{Concept Accuracy}. Unless otherwise stated, results are averaged over three random seeds and reported as mean values. All LLM-based systems use the same title-plus-context prompt template, together with constrained decoding and span-legality checks.

\subsection{Implementation Details}
For LLM-based baselines and our method, we use identical prompts that include the \emph{job title} and \emph{local context}, where the context window size is tuned on the development set. Decoding is constrained to improve boundary validity, and all predictions are post-checked for span legality. Because Gold supervision is limited, we reduce overfitting risk through early stopping on the development set and by cross-checking against Silver annotations. All experiments are conducted on A100 80GB GPUs, and the complete prompts, random seeds, and scripts are released with the scorer for reproducibility.

\begin{table}[t]
\caption{Inter-annotator agreement on a 100-sample gold subset.}
\label{tab:iaa}
\centering
\small
\setlength{\tabcolsep}{6pt}
\renewcommand{\arraystretch}{1.05}
\begin{tabular}{|l|c|c|c|}
\hline
\textbf{Agreement Metric} & \textbf{Precision} & \textbf{Recall} & \textbf{F1 / $\kappa$} \\
\hline
Strict (Exact span)    & 0.707 & 0.426 & 0.532 \\
Relaxed (Overlap span) & 0.829 & 0.500 & 0.624 \\
Token-level Cohen's $\kappa$ & -- & -- & 0.554 \\
\hline
\end{tabular}
\end{table}
\subsection{Main Results on Instruction LLMs}
\label{sec:llm-main}
We compare several instruction-tuned LLMs under the same prompting and decoding protocol, including ChatGPT-4o, Claude, Kimi, DeepSeek, Qwen, and two JobBERT-based baselines. Table~\ref{tab:strict-sf1} summarizes the current strict-setting results.

\begin{table}[!t]
\caption{Strict-setting results (S-F1).}
\label{tab:strict-sf1}
\centering
\small
\begin{adjustbox}{max width=\linewidth}
\begin{tabular}{|l|r|}
\hline
\textbf{Model} & \textbf{S-F1} \\
\hline
ChatGPT  & 0.6700 \\
Claude   & 0.6300 \\
Kimi     & 0.5700 \\  
DeepSeek & 0.5130 \\
Qwen     & 0.2130 \\
JobBERT-skill     & 0.0045 \\
JobBERT-knowledge & 0.0038 \\
\hline
\end{tabular}
\end{adjustbox}
\end{table}

Table~\ref{tab:strict-sf1} shows that proprietary instruction LLMs currently outperform the remaining baselines under the strict span-matching criterion, with ChatGPT obtaining the highest score among the evaluated systems. Across models, we observe that \textbf{L} (language competence) and \textbf{S} (operational skill) are generally easier categories, whereas \textbf{T} (transversal competence) remains more challenging because such spans are often semantically diffuse and behaviorally implicit. The \textbf{K} (knowledge) category benefits from title-plus-context prompting and ontology-aware interpretation, but still exhibits boundary ambiguity in practice.

\subsection{Robustness under OOD and Long-Tail Conditions}
We further evaluate the models under OOD settings to examine robustness to domain and temporal shift. In general, title-plus-context prompting and legality checks help reduce degradation under distribution shift. Among the evaluated LLMs, stronger instruction-tuned models show better stability on long-tail and out-of-domain cases. These observations suggest that contextual prompting and constrained decoding are particularly important for Chinese JobSkillNER, where sparse expressions and domain-specific terminology are common.

\subsection{From Silver to Gold: Class Composition and Confusion Analysis}
\begin{figure}[H]
    \centering
    \includegraphics[width=0.6\textwidth]{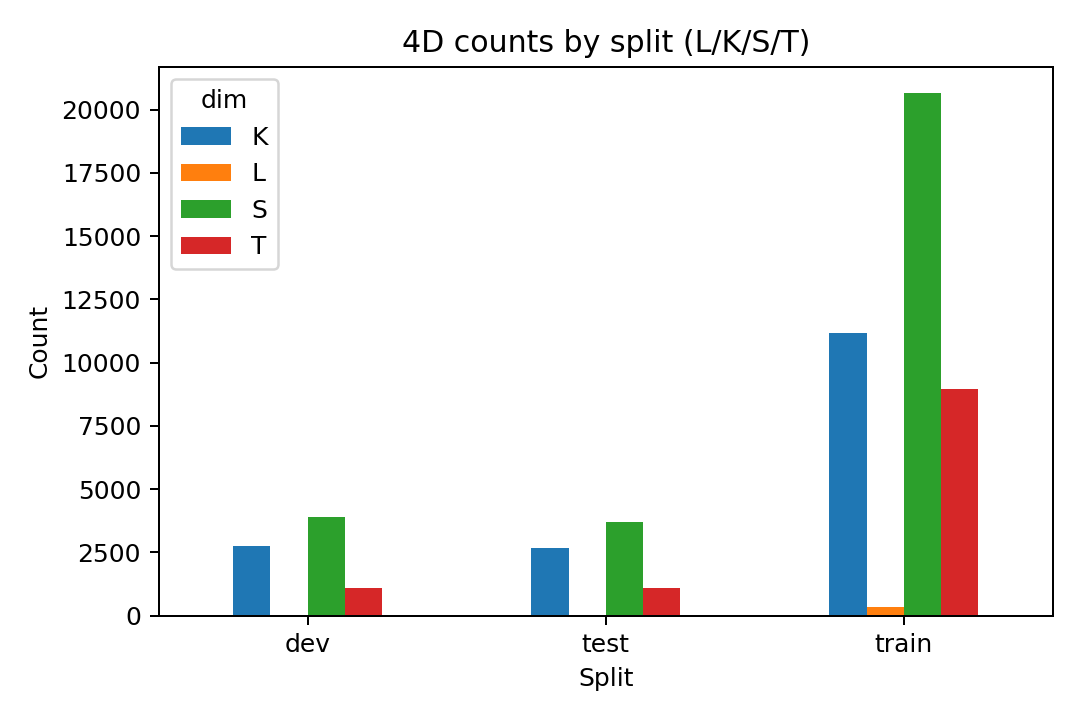}
    \caption{Gold annotations: LSKT distribution (n=200).}
    \label{fig:gold_bar}
\end{figure}

\begin{figure}[H]
    \centering
    \includegraphics[width=0.6\textwidth]{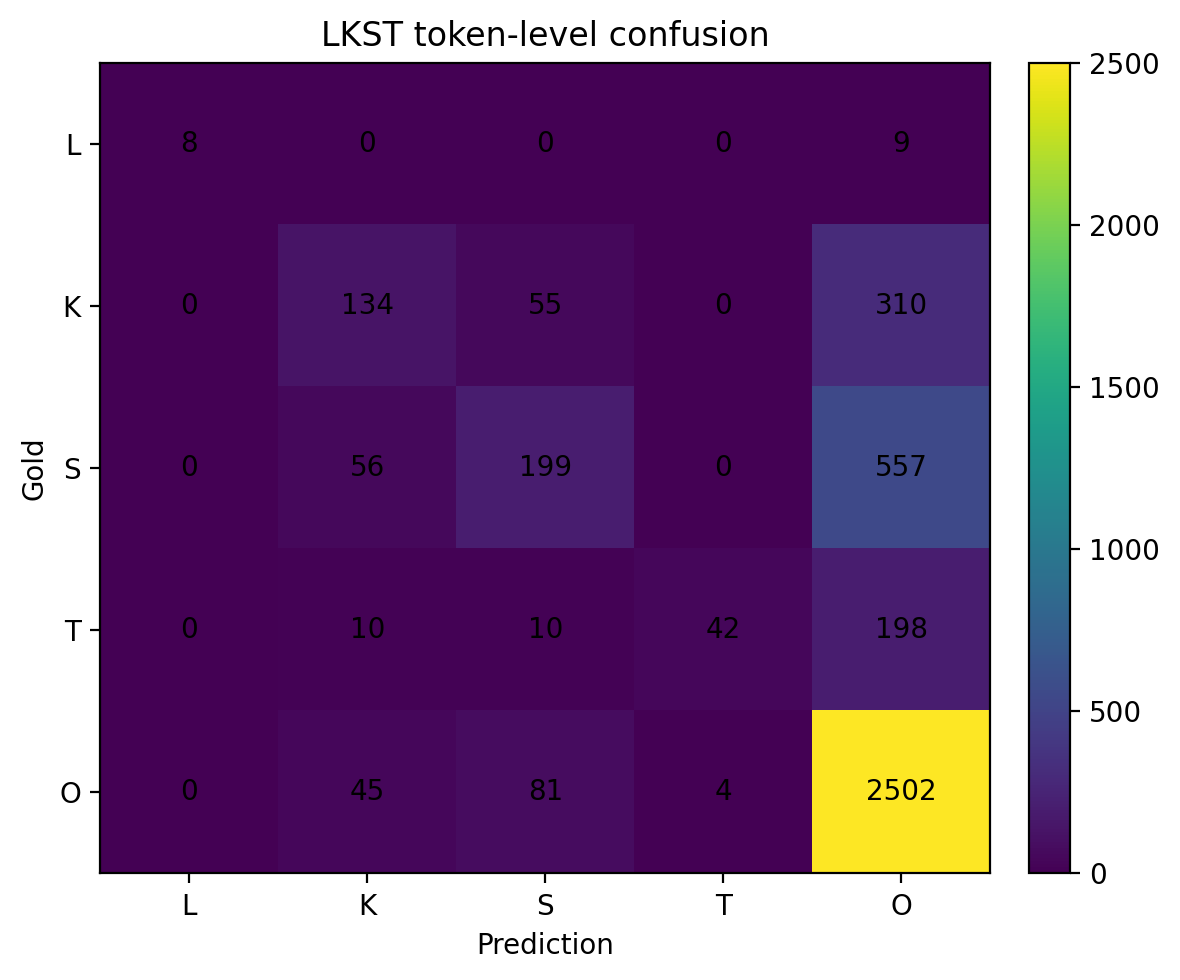}
    \caption{Confusion matrix (Gold vs.\ prediction) under the flat LSKT scheme.}
    \label{fig:cm}
\end{figure}

To better understand annotation outcomes and model behavior, we release 200 \textbf{Gold} items for qualitative and quantitative analysis. These items are human-adjudicated by two coders with one arbiter and are intended for calibration and error analysis. Figure~\ref{fig:gold_bar} presents the class composition under the LSKT scheme, and Figure~\ref{fig:cm} shows the confusion matrix between Gold labels and model predictions under the same flat label space.
The class distribution and confusion matrix provide an initial view of where the current system performs well and where category confusion remains substantial. In particular, confusions are more likely to arise when competency spans are short, context-dependent, or semantically close across categories. These analyses complement the main benchmark results and help characterize the remaining challenges of Chinese JobSkillNER.

\section{Related Work}
\label{sec:related}

\paragraph{Span-level skill extraction and ontology-aligned resources.}
Research on job-skill mining has gradually evolved from lexicon matching and document-level classification to \emph{span-level} extraction with explicit boundary decisions and ontology linkage. In English, \textit{SkillSpan} established expert-annotated hard- and soft-skill spans and demonstrated the value of boundary-sensitive benchmarking for recruitment text mining \cite{zhang2022skillspan}. In Danish, \textit{Kompetencer} further highlighted the importance of ESCO-based supervision and the distinction between boundary detection and concept assignment in job-related competency extraction \cite{jensen2022kompetencer}. Recent surveys have summarized the remaining challenges in labor-market NLP, including ontology alignment, long-tail terminology, and domain shift \cite{senger2024survey}. Despite this progress, there is still no publicly available Chinese benchmark that provides comparable \emph{span-level}, ontology-ready competency annotations for recruitment texts.

\paragraph{Chinese NER benchmarks and domain-specific challenges.}
Chinese NER research has produced a number of influential benchmarks, including social media, newswire, and fine-grained entity datasets, while multilingual evaluations such as MultiCoNER V2 further highlight robustness challenges under noisy and domain-shifted conditions \cite{peng2015weibo,cluener2020,multiconer2023}. Surveys also point out persistent challenges in Chinese NER, such as segmentation ambiguity, character-level granularity, and domain sensitivity \cite{liu2022cnership}. More recent resources extend NER to multimodal or specialized scenarios \cite{cmner2024}. However, these datasets primarily target generic named entities rather than \emph{competency spans} grounded in a structured occupational ontology. As a result, they cannot directly serve as benchmarks for ESCO-aligned job-skill extraction in Chinese recruitment texts.

\paragraph{Modeling paradigms for NER and information extraction.}
From a modeling perspective, recent work has explored both span-based and instruction-based paradigms for entity extraction. Span-oriented methods such as GlobalPointer and SpanMarker have shown strong boundary sensitivity and labeling efficiency compared with BIO-style token tagging, including in Chinese settings \cite{globalpointer,spanmarker}. At the same time, multilingual encoders such as XLM-R provide useful transferability but remain vulnerable to domain mismatch and ontology drift in specialized recruitment data \cite{conneau2020xlmr}. More recently, open-type and instruction-based extraction frameworks, including GLiNER, InstructUIE, and UniversalNER, have demonstrated that LLM-style or distilled instruction-following models can support flexible zero-shot and few-shot extraction across domains and languages \cite{gliner-arxiv,gliner-naacl24,instructuie,universalner,universalner-blog}. These advances suggest that Chinese job-skill extraction should be treated not merely as token classification, but as a boundary-sensitive information extraction task requiring both semantic flexibility and stable type assignment.

\paragraph{Ontologies and evaluation protocols.}
ESCO provides a hierarchical and cross-lingual vocabulary of skills, competences, qualifications, and occupations, and has become an important reference ontology for labor-market intelligence and cross-lingual skill analysis \cite{esco}. Prior span-level job-skill studies emphasize that evaluation should distinguish between at least two aspects: whether a span boundary is correctly identified and whether the extracted span is linked to the correct concept or type \cite{zhang2022skillspan,jensen2022kompetencer}. Motivated by this line of work, we adopt a \textbf{flat, non-nested LSKT} scheme with minimal-sufficient span boundaries, and report \textbf{Exact}/\textbf{Relaxed} Span-F1 together with concept-level evaluation under fixed rules.

\paragraph{LLM-assisted annotation and scalable dataset construction.}
Recent human-in-the-loop annotation pipelines have shown that LLM-generated draft labels, when combined with expert adjudication, can substantially improve annotation efficiency while maintaining transparency and auditability. In such settings, annotation quality is typically monitored using span-level agreement and token-level consistency statistics. Building on this general paradigm, our work adopts a multi-LLM Silver-to-Gold workflow with majority aggregation, confidence-aware drafting, and human adjudication. This design is particularly suitable for Chinese recruitment texts, where span boundaries and competency types often depend on wider contextual interpretation.

\paragraph{Positioning.}
Against this background, we position \textit{Chinese-SkillSpan} as the first Chinese \emph{span-level} competency dataset for recruitment texts that combines three properties: a \textbf{flat LSKT label scheme}, \textbf{minimal-sufficient span boundaries}, and \textbf{offline ESCO-aligned ontology mapping}. In this sense, our dataset complements existing English and Danish resources while providing a reproducible benchmark for Chinese job-ad competency extraction.
\section{Conclusion}
\label{sec:conclusion}

We introduced \textbf{Chinese-SkillSpan}, a Chinese span-level competency dataset for job advertisements aligned with the ESCO framework. The dataset is constructed through an LLM-assisted and human-adjudicated annotation pipeline, and is released together with fixed IID and OOD splits, an open scorer, and supporting documentation for reproducible evaluation. By combining ontology-aware design with Chinese-specific annotation guidelines, Chinese-SkillSpan fills an important resource gap for Chinese JobSkillNER and provides a benchmark for studying competency extraction in recruitment texts.

Our experiments show that the dataset supports effective model training and evaluation, and that reliability-oriented extraction strategies can achieve strong performance under this setting. More broadly, we hope Chinese-SkillSpan can serve as a practical foundation for future research on Chinese job-skill extraction, ontology-aligned information extraction, and data-driven labor-market analysis.

\appendix

\clearpage
\appendix

\captionsetup[figure]{skip=4pt}
\setlength{\textfloatsep}{8pt}
\setlength{\floatsep}{8pt}
\setlength{\intextsep}{8pt}

\section{Supplementary Annotation Materials}

\noindent
Figures~\ref{fig:appendix-a1}--\ref{fig:appendix-a5} summarize the core annotation materials used in Chinese-SkillSpan, including the general rules, concept alignment criteria, conflict-resolution rules, annotation format, and quality-control protocol.

\vspace{0.4em}

\begin{center}
  \includegraphics[
    page=1,
    width=\textwidth,
    height=0.78\textheight,
    keepaspectratio,
    trim=6 8 7 80,
    clip
  ]{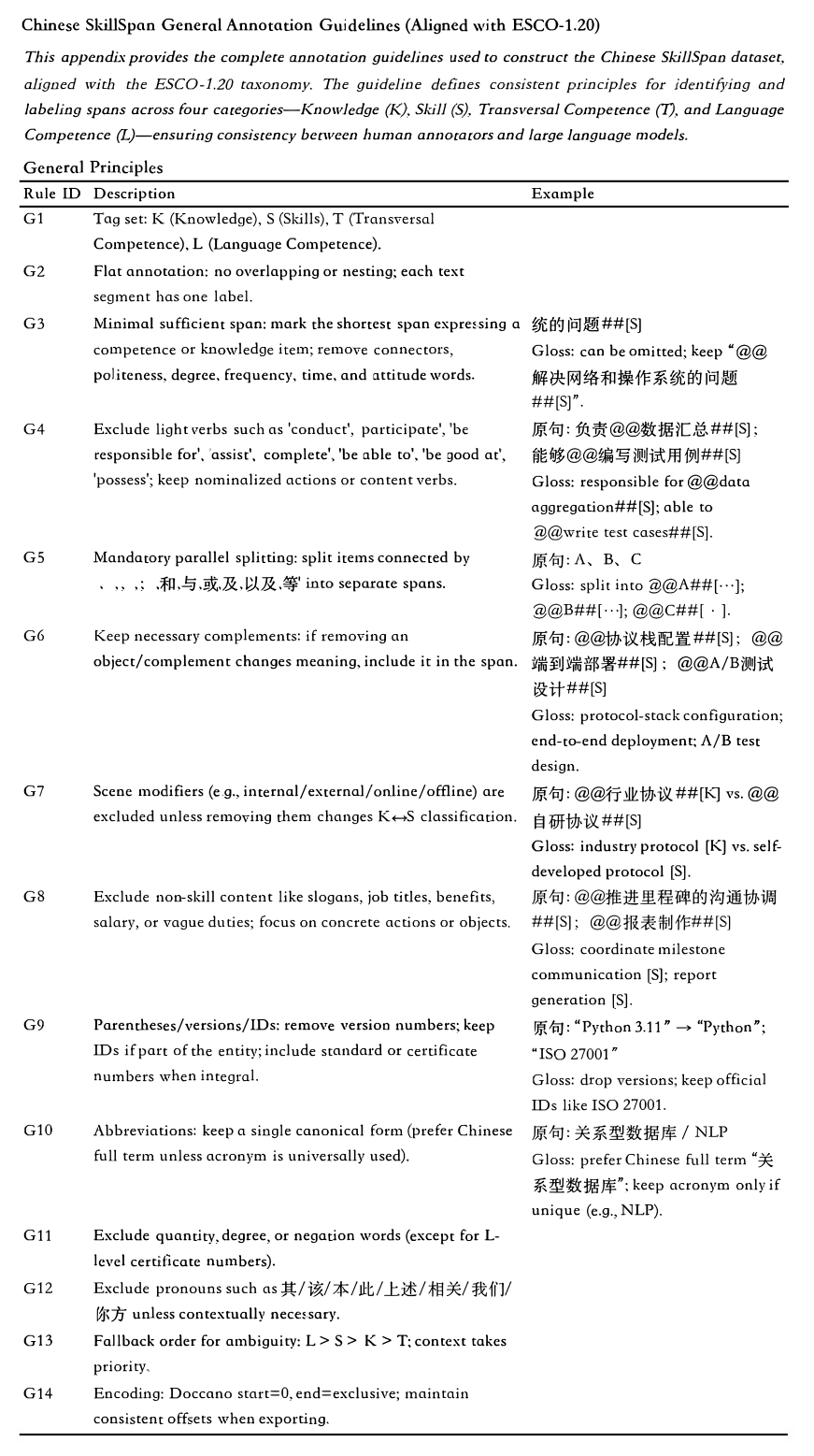}
  \captionsetup{type=figure}
  \captionof{figure}{General annotation guidelines aligned with ESCO-1.20.}
  \label{fig:appendix-a1}
\end{center}

\clearpage
\thispagestyle{plain}
\begin{center}
  \includegraphics[
    page=1,
    width=\textwidth,
    height=0.78\textheight,
    keepaspectratio,
    trim=6 8 6 40,
    clip
  ]{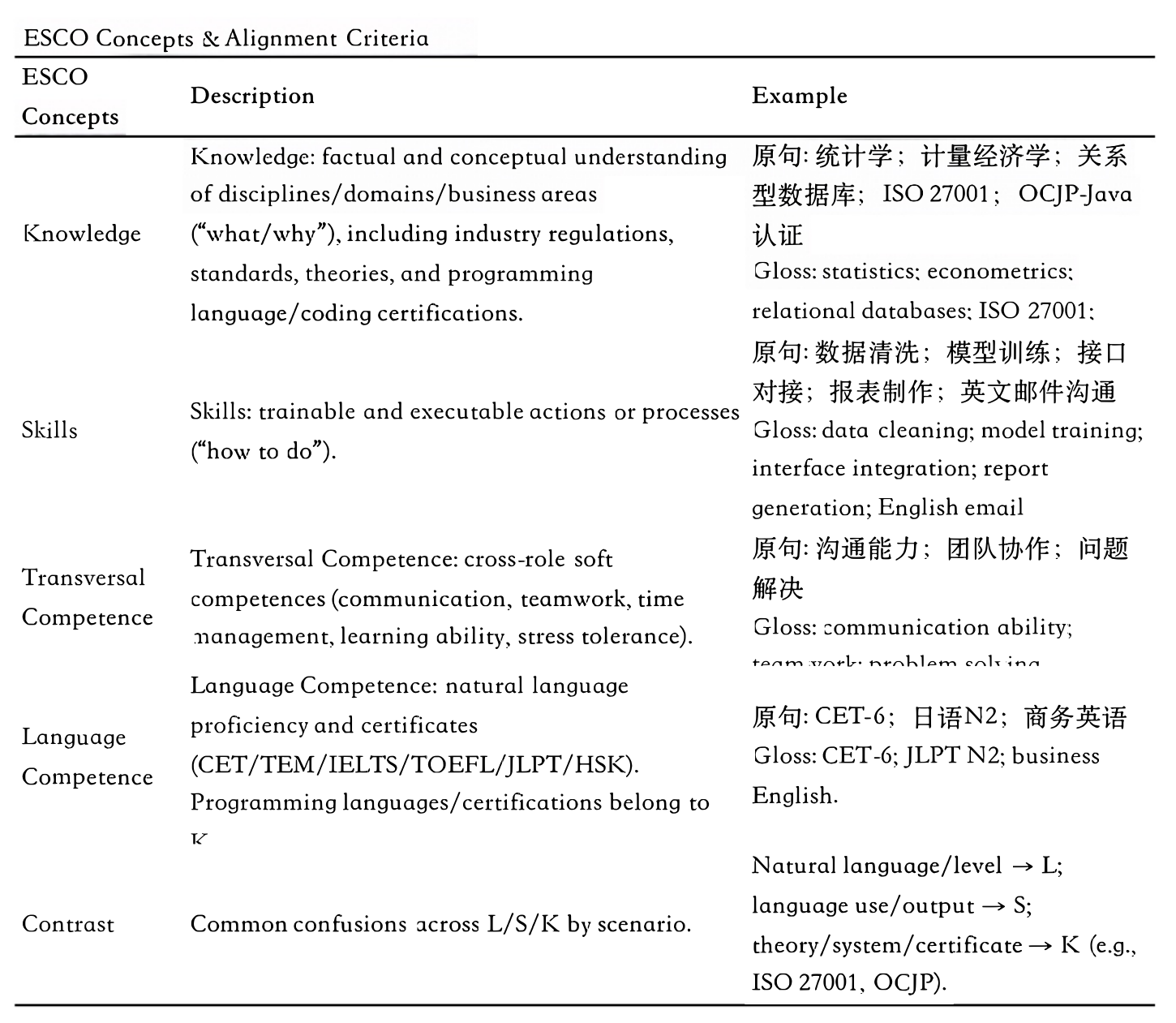}
  \captionsetup{type=figure}
  \captionof{figure}{ESCO concepts and alignment criteria under the LSKT annotation scheme.}
  \label{fig:appendix-a2}
\end{center}

\clearpage
\thispagestyle{plain}
\begin{center}
  \includegraphics[
    page=1,
    width=\textwidth,
    height=0.78\textheight,
    keepaspectratio,
    trim=6 10 6 1,
    clip
  ]{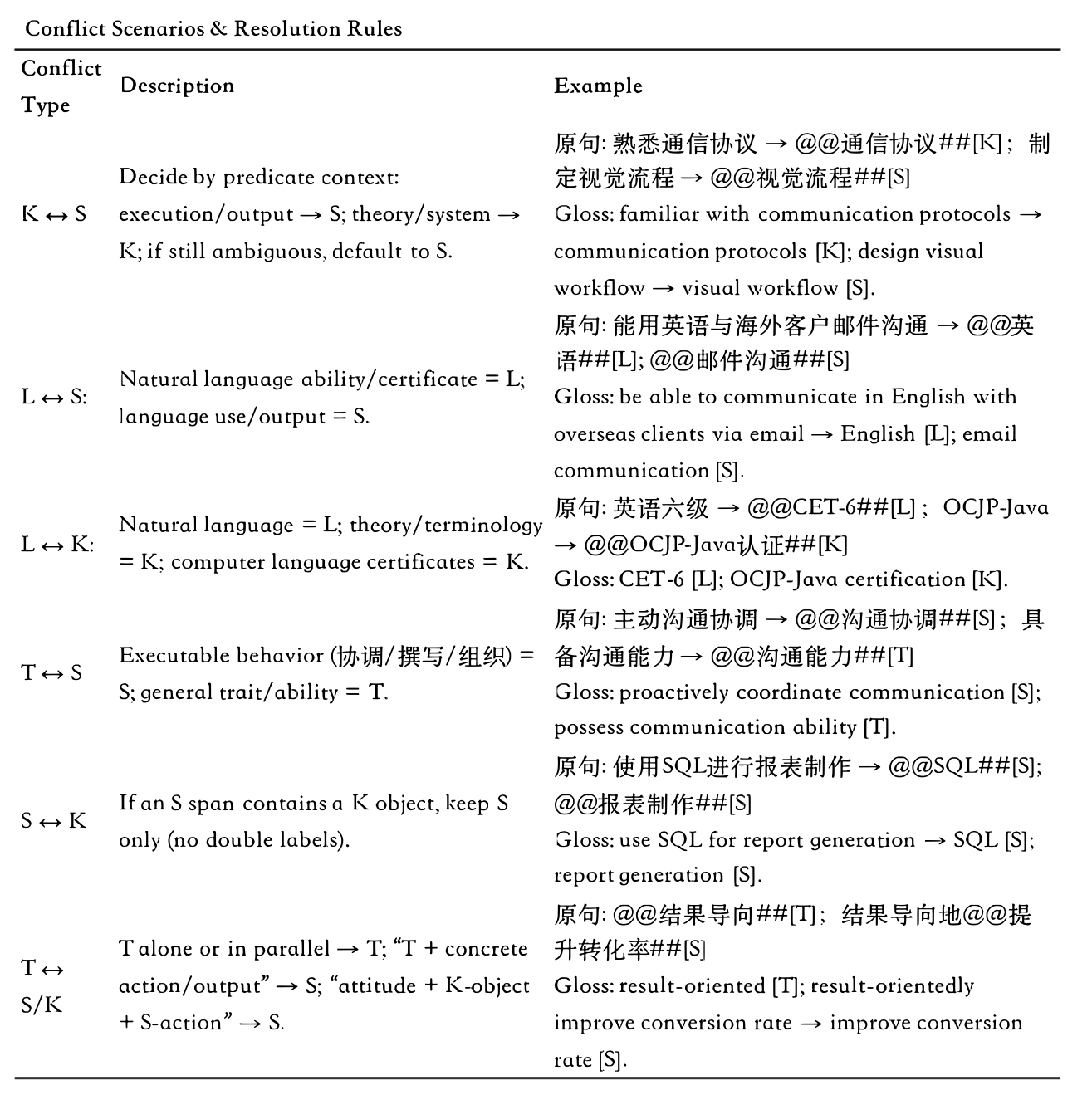}
  \captionsetup{type=figure}
  \captionof{figure}{Conflict scenarios and resolution rules for boundary and category disambiguation.}
  \label{fig:appendix-a3}
\end{center}

\clearpage
\thispagestyle{plain}
\begin{center}
  \includegraphics[
    page=1,
    width=\textwidth,
    height=0.78\textheight,
    keepaspectratio,
    trim=6 10 6 0,
    clip
  ]{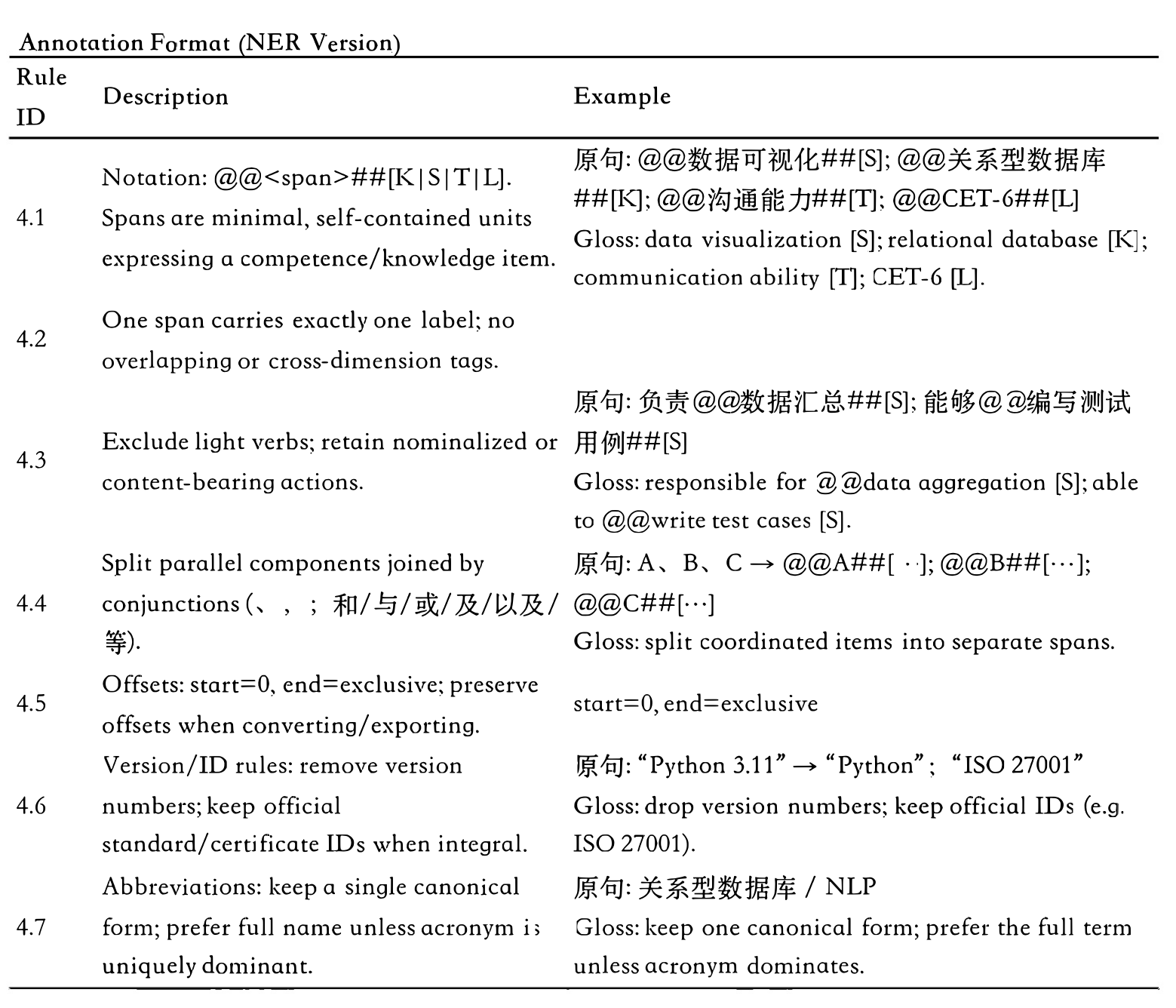}
  \captionsetup{type=figure}
  \captionof{figure}{NER-style annotation format, span notation, and offset conventions.}
  \label{fig:appendix-a4}
\end{center}

\clearpage
\thispagestyle{plain}
\begin{center}
  \includegraphics[
    page=1,
    width=\textwidth,
    height=0.78\textheight,
    keepaspectratio,
    trim=6 10 6 0,
    clip
  ]{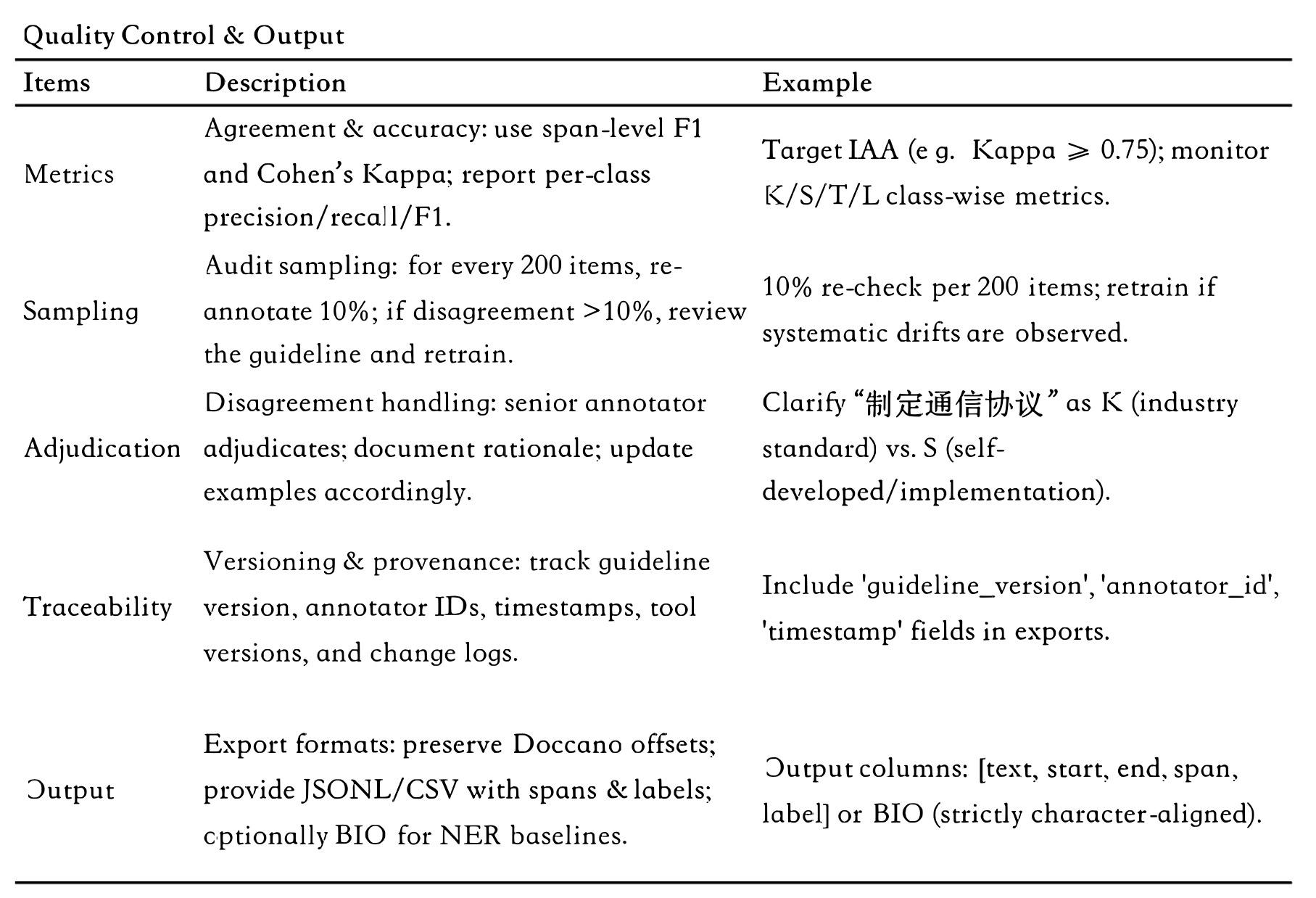}
  \captionsetup{type=figure}
  \captionof{figure}{Quality-control metrics, adjudication rules, traceability, and output conventions.}
  \label{fig:appendix-a5}
\end{center}

\clearpage
\section{Worked Annotation Examples}

\noindent
Figure~\ref{fig:quick-examples} provides representative worked examples under the flat, non-overlapping LSKT scheme. The examples illustrate how span boundaries, label assignment, and category disambiguation are handled in realistic Chinese job-ad sentences.

\vspace{0.4em}

\begin{center}
  \includegraphics[
    page=1,
    width=\textwidth,
    height=0.62\textheight,
    keepaspectratio,
    trim=6 0 6 14,
    clip
  ]{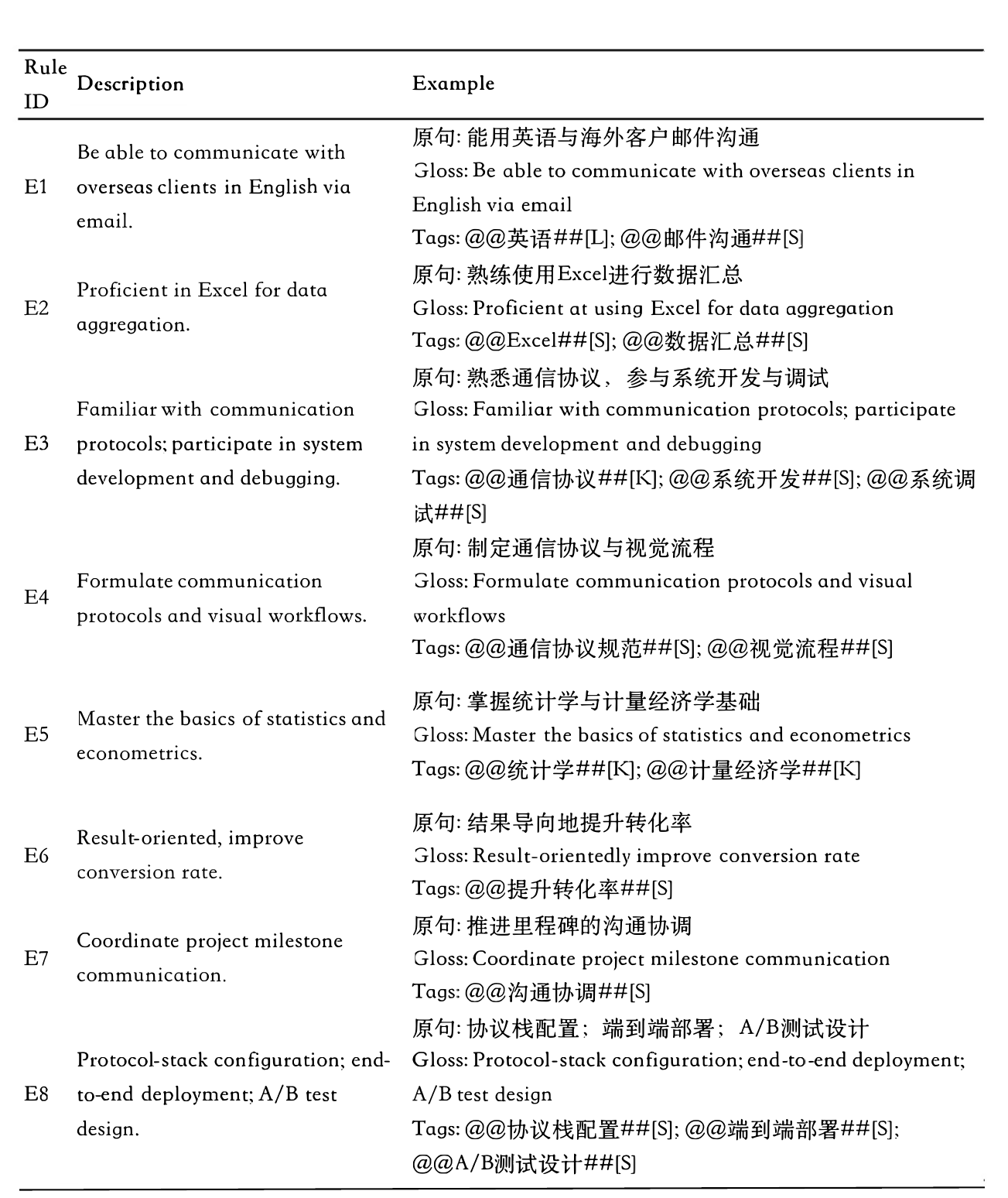}
  \captionsetup{type=figure}
  \captionof{figure}{Worked annotation examples under the LSKT scheme.}
  \label{fig:quick-examples}
\end{center}
\clearpage
\FloatBarrier
\bibliographystyle{splncs04}
\bibliography{9Reference}

\end{document}